\documentclass[twoside,11pt]{article}

\usepackage{jmlr2e}
\usepackage{booktabs}
\DeclareUnicodeCharacter{2212}{-}

\jmlrheading{1}{2000}{1-48}{4/00}{10/00}{Umaid M. Zaffar, Marium Aslam, Imran Malik and Saad Ahmed}

\ShortHeadings{Few-Shot Learning for Biometric Verification}{Zaffar, Aslam, Malik and Ahmed}
\firstpageno{1}

\begin{document}

\title{Few-Shot Learning for Biometric Verification}

\author{\name Saad Bin Ahmed \email sahme532@uwo.ca \\
       \addr Department of Computer Science \\
        Middlesex College, Western University \\
        London, Ontario N6A 5B7, Canada
        \AND
        \name Umaid M.\ Zaffar \email umzaffar@uwaterloo.ca \\
        \addr Department of Data Science\\
            University of Waterloo, 200 University Avenue West \\
            Waterloo, ON, Canada  N2L 3G1
       \AND
        \name Marium Aslam \email maslam.bscs17seecs@seecs.edu.pk \\
        \addr School of Electrical Engineering and Computer Science\\
       National University of Science and Technology\\
       Islamabad, Capital 44000, Pakistan
       \AND
        \name Muhammad Imran Malik \email malik.imran@seecs.edu.pk \\
       \addr School of Electrical Engineering and Computer Science\\
       National University of Science and Technology\\
       Islamabad, Capital 44000, Pakistan}

 \maketitle

\begin{abstract}
In machine learning applications, it is common practice to feed as much information as possible to the model to increase accuracy. However, in cases where training data is scarce, Few-Shot Learning (FSL) approaches aim to build accurate algorithms with limited data. In this paper, we propose a novel lightweight end-to-end architecture for biometric verification that produces competitive results compared to state-of-the-art accuracies through FSL methods. Unlike state-of-the-art deep learning models with dense layers, our shallow network is coupled with a conventional machine learning technique that uses hand-crafted features to verify biometric images from multi-modal sources, including signatures, periocular regions, iris, face, and fingerprints. We introduce a self-estimated threshold that strictly monitors False Acceptance Rate (FAR) while generalizing its results to eliminate user-defined thresholds from ROC curves that may be biased by local data distribution. Our hybrid model benefits from few-shot learning to compensate for the scarcity of data in biometric use-cases. We conducted extensive experiments on commonly used biometric datasets, and our results demonstrate the effectiveness of our proposed solution for biometric verification systems.
\end{abstract}


\section{Introduction}

\begin{figure}
    \centering
    \includegraphics[scale=0.17]{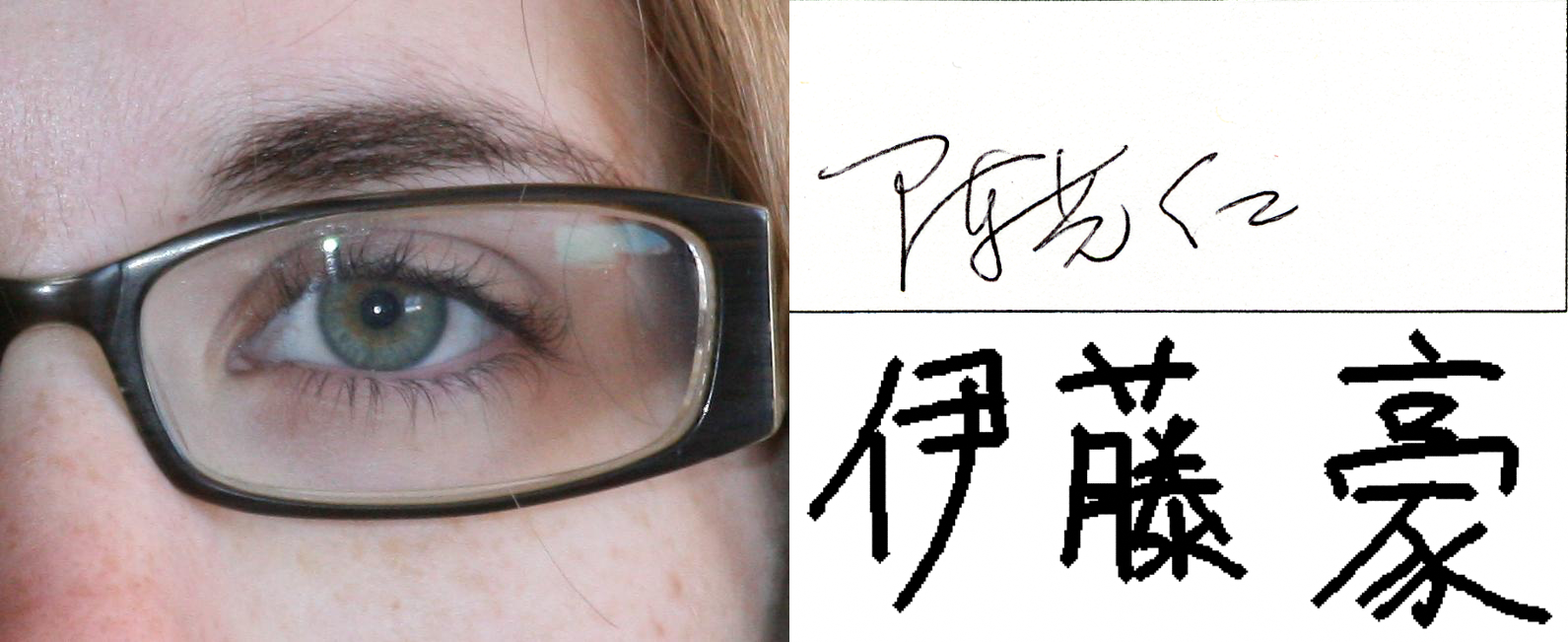}
    \caption{Biometric image samples from UBIPr, SigComp'11 and SigWiComp'13}
    \label{fig:signatures}
\end{figure}

Deep Learning has gained popularity in the past decade, enabling the development of methods for visual recognition problems \citep{krizhevsky2012nips, patil2019i2ct, shubathra2020icesc}. These methods extract features and learn to associate weights with each feature using back-propagation algorithms. Furthermore, such models are commonly used in verification systems that authenticate users for device and network access. For example, handwriting style and facial expressions can be used for user authentication, as depicted in Figure \ref{fig:signatures}.

Biometric verification requires a model that can filter identifying information pertaining to an individual, which necessitates a predetermined threshold indicating the minimum score required to classify a test image as a genuine sample. Receiver Operating Characteristic (ROC) curves \citep{bradley1997elsevier} are commonly used to determine the threshold value where the accuracy of the classification model is highest and the sum of True Positives and True Negatives is maximum. However, biometric verification requires maximum specificity in terms of features to prevent spoofing. The ROC curves are predetermined and computed by iteratively testing the model at different threshold values. In an imbalanced dataset, classification bias can be caused by certain features dominating the samples  \citep{krishnapriya2020tts}. This bias is propagated in the ROC curve, leading to unreliable algorithm performance. In this paper, we propose to replace this mechanism with a self-computed threshold for each input object customized according to its intra-class variation, called the confidence factor. This indicates the confidence that one classifier has relative to another classifier while minimizing the False Acceptance Rate (FAR).

While machine learning algorithms have proven efficient in data-intensive applications, they often struggle when input data is scarce, which poses a particular challenge for biometric verification systems that require high accuracy to detect anomalies.

Meta-learning is a subset of machine learning that teaches itself to learn from transferable knowledge \citep{chelsea2017icml}, and it can be fine-tuned to custom data. One type of meta-learning that has gained prominence in recent years is Few-Shot Learning, which involves predicting and classifying new data with limited training samples in a supervised learning setting. Few-shot learning is a rapidly advancing field that enables the use of sparse data to achieve successful results \citep{wang2020acm}. In this study, we propose an architecture for configuring biometric verification systems that allows for the verification of multiple biometric datasets from different sources in a way that yields competitive results.

In this paper, we aim to make the following contributions.
\begin{enumerate}
    \item We propose a few-shot learning architecture that maintains memory and speed efficiency while achieving fast and reliable results on multiple biometric modalities. This architecture couples a neural network with a boundary-based classifier.
    \item We introduce a novel method for calculating the verification threshold, which involves computing the distance from the decision boundary and applying an inverse sigmoid mapping. This method minimizes the False Acceptance Rate (FAR) and achieves state-of-the-art accuracy.
    \item To compensate for the lack of transparency in neural networks, we use hand-crafted features in conjunction with deep features. This hybrid approach produces efficient and competitive results.
\end{enumerate}


\section{Related Work}
While few-shot deep learning methods for classification have seen substantial progress, data-efficient approaches to deep anomaly detection are still a critical research direction. These anomaly detection schemes, often involving semi-supervised learning, are equipped with mechanisms that detect outliers from a normal class. One-Class Support Vector Machine (OC-SVM) \citep{scholkopf1999nips} maps input data onto a plane and draws a decision boundary, called the hyperplane, to separate genuine and anomalous classes. Support Vector Data Description (SVDD) \citep{tax2004springer} is related to OC-SVM with the objective to find the smallest hypersphere that envelops the data such that radius \(R >  0\) and center C belongs to the feature space of its associated feature mapping. Similarly, Auto encoders were first used for outlier detection by \citep{10.5555/646111.679466} by reconstructing data to create cluster groups, enabling the detection of anomalies. In recent studies, numerous verification-based models have been proposed that conduct verification using deep neural networks \citep{gilles2010jmlr, perera2019tip}. These solutions are performed under different paradigms such as deep hybrid learning \citep{erfani2016elsevier, wu2015iccv}, supervised learning \citep{erfani2017aaai, gornitz2013aiaf} and semi-supervised learning \citep{gilles2010jmlr, perera2019tip}.

Deep Hybrid Learning uses deep learning to extract a set of representative and robust features from unstructured data that are fed into traditional kernel-based anomaly detection techniques, employing a sequential process of extracting features using a neural network and then using machine learning approaches to create a classification model \citep{hodge2018deep}. In this paper, we present a hybrid model that does not rely on this sequential process. Our deep learning model and traditional kernel-based method are, in fact, independent of one another and merged together by an inverse sigmoid mapping of the distance of data points from the decision boundary that serves to estimate the confidence factor for each input data.

Traditionally, hand-crafted features such as Local Binary Pattern \citep{ojala2002ijeas}, Speeded Up Robust Features \citep{zhu2018ijeas} and Local Phase Quantization \citep{ojansivu2008springer}have been used for classification. These features provide users with transparency in decision-making, allowing them to fine-tune their models. However, biometric verification systems require the detection of imposters who can emulate visible features, making it essential to use deep models that can extract representative features. While these models perform well in data-intensive situations, they struggle with small datasets. Some biometric verification algorithms, such as the two-stage Error Weighted Fusion algorithm \citep{Uzair2015PeriocularRP}, require significant manual intervention to remove disoriented data, which can be impractical. Therefore, deep learning methods that are capable of handling sparse datasets are essential for effective biometric verification.

\begin{figure}
    \centering
    \includegraphics[width=\linewidth]{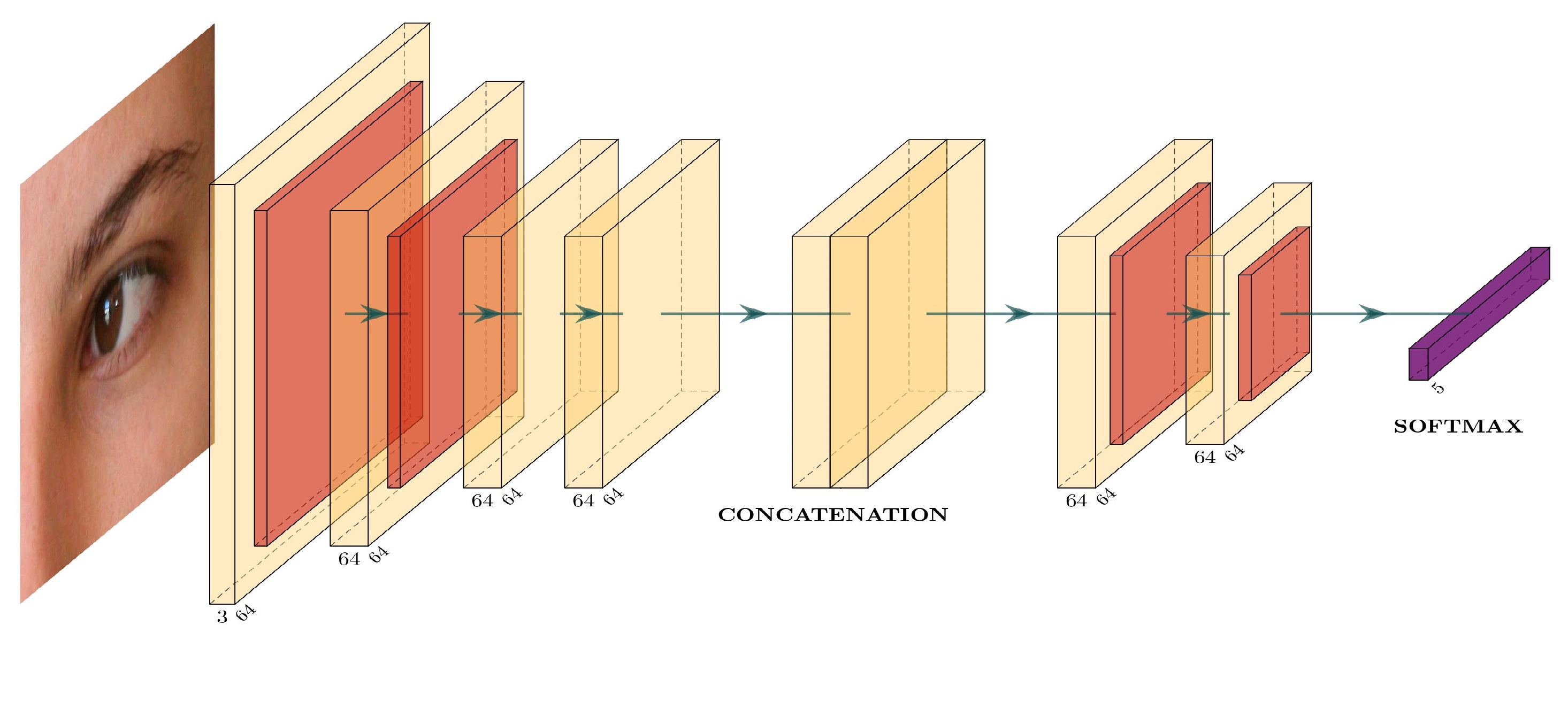}
    \caption{The relation network architecture by \citet{Sung2018LearningTC}. The concatenation layer combines the embedding from the query set and the sample set as both undergo the same process for the first four layers. Following the concatenation layer, the combined feature embedding is forwarded to the next layers.}
    \label{fig:rn_arch_plot}
\end{figure}

The availability of data is a significant issue in biometric verification applications. Recently, many Few-Shot Learning (FSL) models have been proposed for evaluating images under sparse data \citep{FeiFei2006OneshotLO, Lake2011OneSL}. FSL techniques aim to learn a target class T with limited, supervised data by generalizing the model using prior knowledge for task-specific data that follows the same distribution as trained data \citep{wang2020acm}. There are four types of techniques based on prior knowledge:
\begin{enumerate}
    \item \textit{Multi-task learning:} These algorithms learn multiple related tasks simultaneously by exploiting both task-specific and task-generic methods, making them naturally suitable for FSL \citep{Caruana1998MultitaskL}.
    \item \textit{Embedding learning:} TThis method embeds samples from training data into a lower dimension that is less complex, allowing similar data to cluster together while distinguishing dissimilar data conveniently \citep{Jia2014CaffeCA, wang2020acm}.
    \item \textit{Learning with External Memory:} In this method, data from the training set is stored in an external memory in key-value format, and each test sample represents the contents of the memory as a weighted average \citep{Miller2016KeyValueMN, Sukhbaatar2015EndToEndMN}.
    \item \textit{Generative Modeling:} This method estimates the probability distribution \( p(x)\) from the observed \(x_i\)'s, where i is the index of observed data, using prior knowledge and involving the estimation of \(p(y|x)\) and \(p(y)\) \citep{Rezende2016OneShotGI}.
\end{enumerate}

The proposed model builds upon the work of \citet{Sung2018LearningTC}, which uses embedding learning to train a two-branched Relation Network (RN) that compares query images to few-shot labelled sample images. In this paper, we introduce a modified version of the RN architecture for biometric verification. Our model compares query images to a set of K-shot samples, using only genuine images in the sample set. To minimize biases within classes, as explained by \citet{Sung2018LearningTC} we inhibit the element-wise summation of embedding modules, as depicted in Figure \ref{fig:rn_arch_plot}. We derive the similarity score threshold using a confidence factor extracted from kernel-based estimation. Our simple approach produces competitive results and offers significant improvements in FAR compared to previous approaches, providing a single model solution to multi-modal biometric sources.

\begin{figure*}
    \centering
    \includegraphics[width=\linewidth]{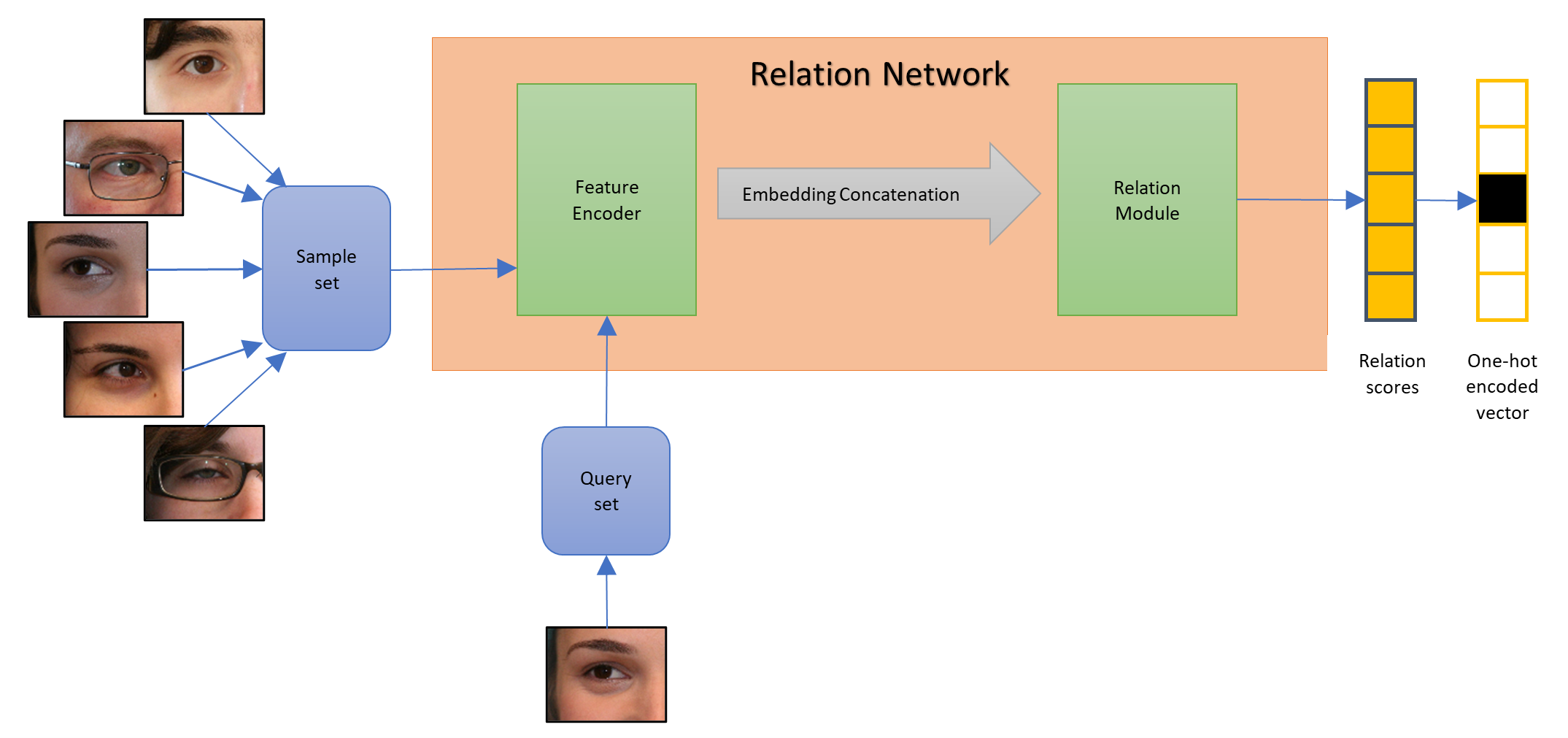}
    \caption{The relation network architecture used in proposed biometric verification system. The sample set consists of images that are genuine and the query set image is being questioned for its authenticity. After comparing the query image with sample sets, the embedding that yields highest score is the most similar image through deep feature comparison.}
    \label{fig:rn_architecture}
\end{figure*}
\section{Problem Definition}
For the purpose of biometric verification, we are examining the task of few-shot learning. In this section, the proposed system is divided into two parts: a few-shot learning model and a threshold computing mechanism for authentication. We employ few-shot learning algorithm that uses Relation Networks which are also required to have a training set, support set and a test set. The support set and test set have the same label space and the support set contains $K$ labelled examples for each of $C$ unique classes, , calling the target few-shot problem a $C$-way-$K$-shot problem.

In principle, the support set can be used to train the classifier but because of the limited label space, the performance is, usually, not satisfactory. Therefore, we perform meta-learning on training set, which has its own label space disjoint from the support set, in order to extract transferable knowledge. The training set consists of \(N_i^g\) number of images from each class \(i\) from the genuine images \(g\). In each class, there are genuine images that are anomalous for other classes. In the meta-learning phase, this set is used to train the relation network that learns to compare a pair of two images. We emulate the episodic training setting proposed by \citet{Vinyals2016MatchingNF} to benefit from our complete training set where, in every episode random set of sample and query images are used to maximize generalization and test accuracy. The sample set $S$ contains $K$ reference images from each of the selected $C$ classes, from the $G$ genuine classes, \(S = \{(x)\}^m\) where \(m = K\times C\) and $x$ represents a tuple of sample images and labels. The  query set contains $N$ number of images from the same $C$ classes, \(Q = \{(y)\}^n\) where \(n = N\times C\) and y represents the tuple of query images and label. All \(n\times m\) images are encoded by the feature encoder and produce \(n\times m\) embeddings. Each query embedding, \(f(y_i)\), is batched with each of the $C$ sample embeddings creating \(n\times C\) pairs per episode. Considering a single batch, with respect to a query embedding, we would have $C$ pairs of embeddings in which the embedding \(f(y_i)\) is concatenated with $C$ sample embeddings. Out of these $C$ pairs, only one is the correct genuine pair, while the others are anomalous. We opt for a modified version of 5-way-1-shot learning setting as illustrated in figure \ref{fig:rn_architecture} for our model training to provide for the possibility of forgeries in our dataset. For threshold computation, we devise an independent data-preprocessing pipeline to extract hand-crafted features and utilize them to perform one-class classification using the One-Class SVM on the test feature vector against the target class. The distance from the hyperplane to the test sample is further processed using a scaling function and exploited via an Inverse Sigmoid Mapping to extract the confidence factor based on the distance. This factor is used as a threshold value for the few-shot RN relation score. The detailed pipeline is described in Section \ref{section:model}.

\section{Model}
\label{section:model}
\begin{figure*}
    \centering
    \includegraphics[width=\linewidth]{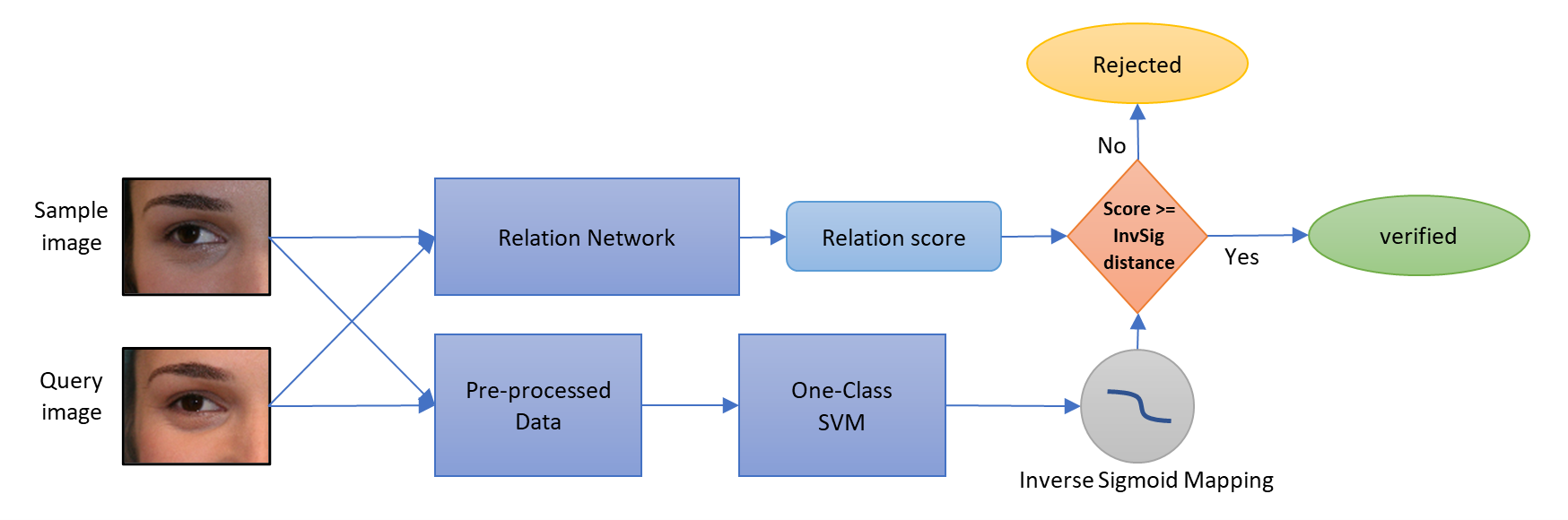}
    \caption{End to end architecture for biometric verification through our approach. This approach encapsulates OC-SVM and a shallow neural network to produce a fast, lightweight solution that can produce results that can be compared to fully connected networks. The hand-picked features allows us to visualize the feature that are being used for initial prediction. }
    \label{fig:fusion_archit}
\end{figure*}

The biometric verification architecture is constructed around two networks, as illustrated in Figure \ref{fig:fusion_archit}: OC-SVM that extracts a confidence factor using hand-crafted features and another Relation Network that learns to compare images based on their deep embeddings. Each task is represented by an image from a unimodal source that is sub-divided into genuine and imposter set that constitutes \(N^g\) and \(N^f\) images. After a series of preprocessing steps extracting hand-crafted feature descriptors from an image \(g_i\) is fed to OC-SVM. For each \(g_i\) that is mapped on the plane, the distance \(d_i\) from the hyperplane is measured which is input to a sigmoid function \(\sigma(d_i)\). We subtract one from this value to calculate the inverse mapping \(O_i\) which is used to compute the confidence factor, 
\begin{equation}
    O_i = 1 - \sigma(d_i), i = 1, 2, 3, ..., N^g 
\end{equation} 
A numerical value is assigned to the OC-SVM decision based on this output. Using this threshold value as a level against the relational score from RN allows us to maximize our accuracy while minimizing FAR. A detailed overview of all modules involved in this process is as follows,

\begin{figure*}
    \centering
    \includegraphics[width=\linewidth]{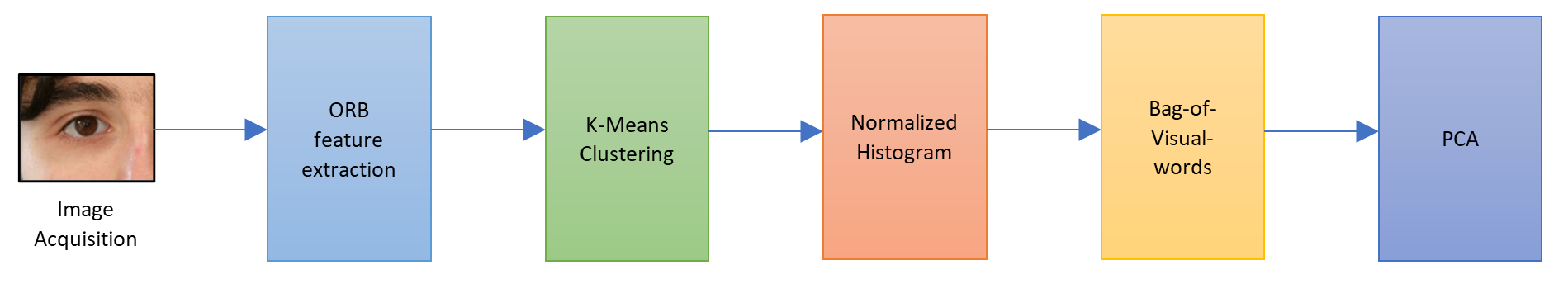}
    \caption{The data preprocessing pipeline is used for all modes of biometric datasets. These set of techniques allows us to quantize the hand-picked feature so that we can produce our OC-SVM model.}
    \label{fig:dp_archit}
\end{figure*}

\subsection{Preprocessing}
\(N^g\) images compose the target class while \(N^f\) images constitute the rest of the classes augmented as an anomalous class. We use OC-SVM, a class boundary based classification technique, to train our kernel based model. Therefore, only \(N^g\) images needs to be preprocessed. To efficiently extract features from the images, Oriented FAST and Rotated Brief \citep{Rublee2011ORBAE} are applied. For each image \(g_i\), the number of local feature key points vary under different environmental conditions such as light, orientation, angle and shadow. In order to resolve this inconsistency, we employ a bag of visual words to constrain the number of feature descriptors for each image to a set number of visual words. This is achieved by using a K-Means clustering algorithm trained on each raw feature descriptor of every image \(g_i\). \(N^g\) images are fed to K-Means Cluster and a normalized histogram is created that represents the occurrences of key points in each cluster of an image. We store this preprocessed data for each image that represents a bag-of-visual-words. This technique runs a risk of having empty clusters that can spoil the dataset. Therefore, a Principle Component Analysis (PCA) followed by this technique with an intention to reduce the dimensions of feature descriptor. The sequential steps are outlined in Figure \ref{fig:dp_archit}.

\subsection{One-Class SVM}
The previously determined feature vectors for our genuine set is input to OC-SVM \cite{scholkopf1999nips} to develop a model that separates the genuine images from the origin so that the distance between the hyperplane and origin is maximized. In our work, the distance between the decision boundary and the data points plays a significant role as a confidence inducer. Using this model, a sample image can either be classified as genuine or anomalous based on its position relative to the decision boundary. In this initial estimation, positive distances and negative distances are calculated to represent genuine and anomalous class respectively. The sigmoid function is used to infer the confidence factor for the test image as represented in equation (1). The distance \(d_i\) is mapped to a range of  \([0,1]\) as an output of sigmoid function. Ideally, according to this function, the test samples classified as anomalous are mapped to \([0,0.5]\) and genuine samples to \([0.5,1]\).

\begin{figure*}
    \centering
    \includegraphics[width=\linewidth]{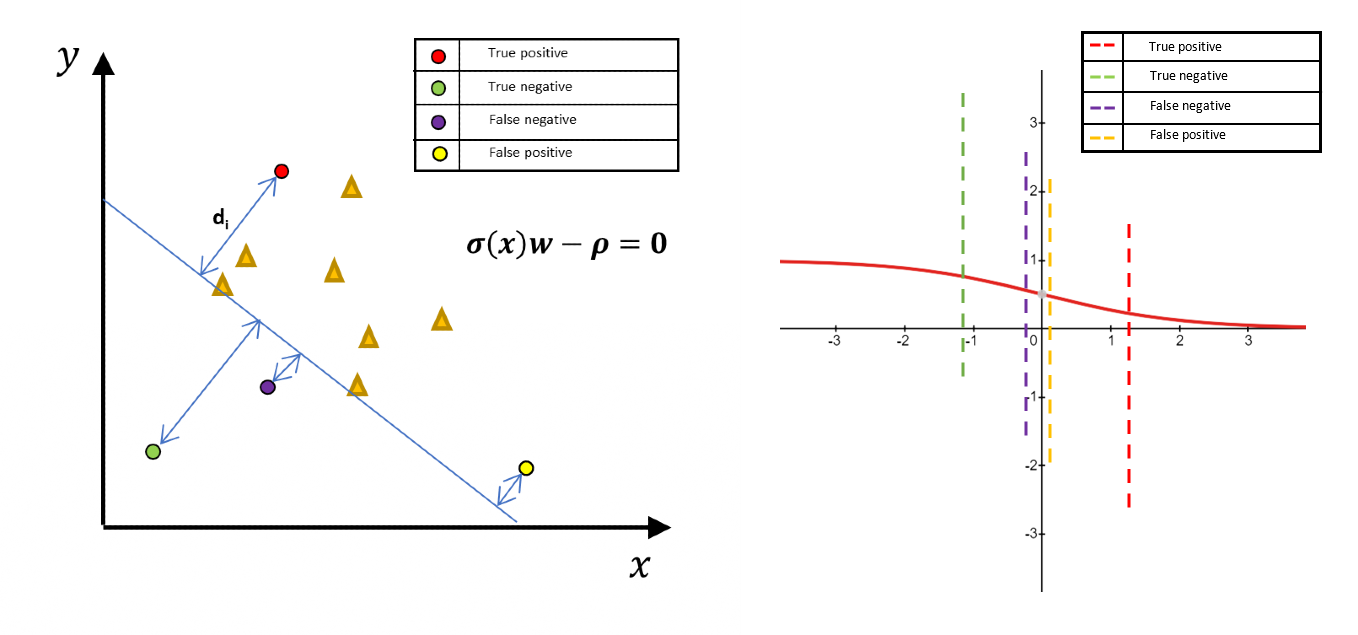}
    \caption{Data points as they are mapped on a 2-D plane and classified according to their distance from the decision boundary which is then mapped using inverse sigmoid function that gives us thresholds as shown}
    \label{fig:types}
\end{figure*}

The probability of misclassification increases if the test sample lies close to the decision boundary depending on intra-class variation. Since these samples have a similar risk of misclassification, their inverse sigmoid output \(O_i\) is estimated close to 0.5. This confidence factor will give an equal chance to the relation network to compute the relational score that will determine the final verdict regarding the distribution of sample data. The inverse sigmoid mapping is explained under four scenarios that correspond to OC-SVM and how any data that is initially mislabelled can be amended. An ideal scenario is illustrated in Figure \ref{fig:types} where such data points are processed using OC-SVM to show the concept of using the distance to compute the confidence factor.

\subsubsection{True Positives (TP)}
Test samples that are evidently similar to the genuine set based on visual observation and handcrafted features are likely to be mapped furthest from the hyperplane towards the positive direction, yielding a confidence factor of \(0.5 << \sigma(d_i) < 1\). This indicates that the model has inferred a high probability for the sample test, thereby placing confidence in its result as a genuine class. When the sigmoid function is inverted, it will result in an output of \(0 < O_i << 0.5\). , which represents a low threshold for the relational score. This is because the sample is already evidently similar to the genuine distribution, and does not require thorough confirmation via deep feature comparison to classify it otherwise. However, due to high intra-class variation, the sample may lie near the decision boundary of the OC-SVM. In this scenario, the \(O_i\) will yield an output close to 0.5, and the relation network will play its role in conducting a deep feature comparison of embeddings that will yield a result greater than this threshold.

\subsubsection{True Negatives (TN)}
If the test samples are correctly classified by the OC-SVM as anomalous class, then the distance from the decision boundary will be in the negative direction. If the sample image is evidently dissimilar to the genuine set, the sigmoid function will give \(0 < \sigma(d_i) << 0.5\), and likewise, \(0.5 << O_i < 1\). This implies that the sample image will require a high relational score based on deep feature comparison to overcome its classification as an anomalous class. As in the case of TP, the sample image can also lie near the decision boundary if there is high intra-class variation in the genuine set. In that case, the relation network will output a relational score less than \(O_i\) to confirm that the sample image is anomalous.

\subsubsection{False Positives (FP)}
The test samples that exhibit some similarity to the images in the genuine set, but do not correspond to its distribution, will lie near the decision boundary as they are neither completely similar nor dissimilar to the genuine set. In this case, the sigmoid function will output \(0.5 < \sigma(d_i) << 1\), and corresponding, \(0 << O_i < 0.5\). Therefore, the relation network will be responsible for comparing embeddings to give a relational score less than the confidence factor, and correct the mislabeled sample image.

\subsubsection{False Negatives (FN)}
The test samples are slightly dissimilar to the distribution of the genuine set; however, they do belong to the genuine class. For example, a tired person's signature may be slightly different compared to when the person was alert. This sample also lies near the decision boundary towards the negative direction with \(0 << \sigma(d_i) < 0.5\) and \(0.5 < O_i << 1.\) TThe relation network will output a relational score greater than \(O_i\),, correcting the earlier label of this sample.

\subsection{Relation Network}

The Relation Network constitutes the second module of our architecture, which is responsible for conducting few-shot learning and relation score estimation for biometric verification, as previously mentioned. The CNN architecture comprises two sub-networks: the Feature Encoding Module and the Relation Module. The Feature Encoder generates embeddings of each image independently, and the resultant class sample and query sample embeddings are concatenated to form sample-query pairs. These pairs are fed into the Relation Module, which outputs the relation score of each pair of embeddings.
We can define each episodic batch as a set that consists of one genuine sample image of the target class, four random anomalous class samples, and a query image belonging to the target class. The network not only learns to discover similarities between genuine class samples but also to determine differences between the query sample and anomalous classes. The output relation scores, ranging from 0 and 1, describe how similar the query image is to the genuine class sample and how different it is from the anomalous class samples.
Mean Squared Error (MSE) is calculated, where the genuine score is compared with 1 and the anomalous scores with 0. This error is back-propagated as the loss function of the network, updating the corresponding weights of the Feature Encoder and the Relation Module sub-networks. In multi-class datasets with no forgery data, the training is conducted similarly to \citet{Sung2018LearningTC}, where the network learns to compare and differentiate between images of distinct classes. During testing, each target class is compared with either a genuine, i.e., target class random sample, or an anomalous sample, i.e., from any class other than the target class. The network outputs a relation score of the sample with the target class sample.
In datasets that contain genuine and forged samples of each class, the method is altered. First, we train the network in the same multi-class procedure mentioned above. Then, we further fine-tune the model by making it learn to differentiate between genuine and forged samples of each target class. The sample set contains genuine target class images, while the query set has either a forged or genuine image from the combined forgery dataset. The network outputs a relation score, and if the ground truth value of the query is genuine, the error is computed between the score 0 and 1. Similarly, if the ground truth is forged, the error is measured with 0. This error is back-propagated as the loss in the unfrozen layers of the network.
For testing, the process is similar, where the support set contains genuine samples, and the test set contains both forged and genuine samples that are to be tested. The network is cross-validated after every 10 episodes, where the weights yielding the highest accuracy are saved.

\section{Network Architecture}

The proposed architecture combines a few-shot learning model with a feature extraction module for a kernel-based, one-class classification technique. The encoding module in the Relation Network consists of four convolutional blocks, which is similar to most few-shot embedding modules, while the relation module consists of two convolutional blocks. As designed by \citep{Sung2018LearningTC}, each block includes a 64-channel \(3\times3\) kernel convolution, batch normalization, and ReLU activation. The first two blocks of each module also contain a \(2\times2\) maxpooling layer. The blocks are followed by two fully-connected linear layers with dimensions 8 and 1, respectively, with the last uni-dimensional layer corresponding to the final output, that is, the relation score for each query-sample pair. The output vector also passes through a sigmoid function to ensure the scores lie in the range of (0,1). The data-preprocessing timeline, as described above, is followed to create feature vectors for the entire dataset before evaluation. During the testing phase, while the Relation Network computes the relation score of the query image, the feature vectors of the target class are fed into the OC-SVM to create a decision boundary. The distance from this boundary to the test sample feature vector, as determined by the OC-SVM, is scaled and mapped through an inverse sigmoid function that outputs a confidence factor. This factor is realised as a threshold value for the Relation Network output score. The network architecture is summarized.
 
\section{Experiments}
The most prominent types of biometrics include signatures and periocular region analysis. We will evaluate our technique in the aforementioned domains through the following datasets: UBIPr  \citep{Padole2012PeriocularRA}, SigComp 2011  \citep{Liwicki2011SignatureVC}, and SigWiComp 2013  \citep{malik2013icdar}

In our experiments, the majority of absolute distances computed by the OC-SVM appear to be concentrated between 0 and 0.3. Most values are very small and close together, and can be used as inputs to our inverse sigmoid function. As a result, the sigmoid function produces approximately the same threshold value of approximately 0.5 most of the time. In this instance, user-specific thresholding serves no purpose. To overcome this, we have scaled our OC-SVM distances using a piecewise function.
This function has two parts: for inputs greater than |0.98|, it returns the same value, while inputs between [-0.98,0.98] are fed into a scaled Tanh function. Furthermore, we have included a multiplying factor of approximately (\(\sim\)2.36) in the Tanh function to increase the range of output values. This multiplying factor also ensures that both parts generate equal outputs when the inner limit values (\(\pm\) 0.98) are input, resulting in a smoother transition between the two parts. The output of this function is considered as an input to the confidence-mapping inverse sigmoid function.

\subsection{UBIPr}
This dataset is a version of UBIRIS.v2 where the images are cropped in a way that cover a wider part of the ocular region \citep{tax2004springer}. This dataset contains 10,252 RGB images captured by Canon EOS 5D Camera from 344 subjects with $(501\times401)$ pixels of resolution. Images in this dataset are taken from a distance of 4m to 8m to include distance variability. Due to gaze and pose variation between $(0^\circ, 30^\circ, −30^\circ)$ and hair covering the periocular region, this dataset suffers from the problem of occlusion. One subject contained only a single image so we removed that from our experiments.

\subsubsection{Training}
\label{experiments:ubipr}
All input images were resized to (\(128\times128\)) and normalized according to the ImageNet statistics. The proposed architecture requires the combination of two independent models: OC-SVM and Relation Network. The Relation Network is trained episodically using 5 classes at a time (5-way), 1 sample image per class (1-shot), and 25 query images; 5 for each class. This results in a total of 30 embeddings per episode and 25 concatenated pairs being fed into the relation network module, over 10,000 training episodes. For our data preprocessing, we set the number of clusters to 100 during the development of the bag-of-visual-words. The normalized histogram undergoes Principal Component Analysis (PCA) to reduce redundant dimensions. This dataset of extracted features is made ready before testing. The vector belonging to the target class is fed to the OC-SVM to create a decision boundary that encapsulates the known distribution.

\subsubsection{Results}
We combine our models to create a few-shot learning verification architecture. Emulating \citet{yilmaz2011ijcb}, we determine the accuracy by averaging over 200 randomly generated episodes from the test set with 1 query image batched per target class. We conduct testing by computing the relation or similarity score of a random test sample with the genuine class sample. The randomly selected test sample has an equal chance of being genuine or anomalous, where all classes apart from the genuine target class are grouped together as anomalous. We evaluate our model’s performance using two separate experiments: in one, we compute relations between a single target class sample and the query image, and in another, we assess the maximum relation score between the query image and 5 samples per target class. This improves our accuracy via more comparisons, reducing the impediment caused by occlusion. The results achieved using this method are listed in Table \ref{tab:ubipr}, along with previous methods for periocular image verification.

\begin{table}[h!]
\caption{UBIPr Results Comparison}
\setlength{\tabcolsep}{3pt}
    \centering
    \begin{tabular}{c c c c c}
    \toprule
    \textbf{Model} & \textbf{Position} & \textbf{Accuracy} & \textbf{FAR} & \textbf{EER} \\ \midrule
    Discriminitive CF \citep{smereka2016} & \begin{tabular}{c} Left \\ Right\end{tabular} & \begin{tabular}{c} 78.59\% \\ 84.14\%\end{tabular} & \begin{tabular}{c} 0.001\% \\ 0.001\%\end{tabular} & \begin{tabular}{c} 8.00\% \\ 6.46\%\end{tabular} \\ \midrule
    UBIPr CF\citep{smereka2017} & \begin{tabular}{c} Left \\ Right\end{tabular} & \begin{tabular}{c} 77.95\% \\ 84.33\%\end{tabular} & \begin{tabular}{c} 0.001\% \\ 0.001\%\end{tabular} & \begin{tabular}{c} 8.93\% \\ 7.27\%\end{tabular} \\ \midrule
    PPDM \citep{smereka2015} & \begin{tabular}{c} Left \\ Right\end{tabular} & \begin{tabular}{c} 81.49\% \\ 85.28\%\end{tabular} & \begin{tabular}{c} 0.001\% \\ 0.001\%\end{tabular} & \begin{tabular}{c} 7.67\% \\ 7.09\%\end{tabular} \\ \midrule
    Proposed Model & \begin{tabular}{c} Left \\ Right\end{tabular} & \begin{tabular}{c} \textbf{91.01\%} \\ \textbf{90.09\%}\end{tabular} & \begin{tabular}{c} \textbf{0.001\%} \\ \textbf{0.001\%}\end{tabular} & \begin{tabular}{c} \textbf{4.49\%} \\ \textbf{4.5\%}\end{tabular} \\ \midrule
    
    \end{tabular}
    \label{tab:ubipr}
\end{table}

\subsection{SigComp 2011}
The SigComp dataset contains both simultaneous online and off-line signature images, but we only use off-line images for our purpose. The collection is divided into two separate groups: Dutch and Chinese signatures. For the Dutch set, the training set contains 362 offline signatures of 10 reference writers, with 235 genuine signatures and 123 forgeries of these writers. The test set contains a total of 1932 offline signatures of 54 reference writers, including forgeries of them. The reference set contains 646 genuine signatures, while 1287 questioned signatures are provided for testing.
The Chinese dataset has a training set of 575 offline signatures of 10 reference writers and their forgeries, containing 235 genuine signatures and 340 forgeries. The test set for this subset contains 602 offline signatures of 10 reference writers and their respective forgeries, with 115 referenced and 487 questioned signatures. The images were collected using a WACOM Intuos3 A3 Wide USB Pen Tablet and MovAlyzer software \citep{Uzair2015PeriocularRP}. All input images are resized and cropped to (\(128\times128\))and converted to 1 input channel.

\subsubsection{Training}
The data preprocessing and OC-SVM procedure are the same as mentioned in Section \ref{experiments:ubipr}. However, the Relation Network training method is modified. First, we train the Relation Network using 5-way-5-shot with four query images per class for classifying between genuine signatures of different subjects for 10,000 training episodes. Then, we further fine-tune the model over 1,000 episodes under one-class settings by learning to compare genuine and forged samples of the same subject. In this approach, a query image is batched with a target class sample, where the genuine query would have a ground truth value of ‘1’, and a forgery would be classified as ‘-1’. This enables the model to capture and comprehend the fine-grained differences in details between a signature and its forgery.

\subsubsection{Results}
We evaluate the proposed architecture on the provided test set for each subset. Testing is conducted by using the provided test reference set as a support set and the provided questioned set, containing genuine and forged signatures, as the test set. Each questioned image of a target class is compared with either 1 or 5 genuine samples of the class under separate experiments. The threshold-determining process works similarly, and the results of the fused architecture are summarized in Tables \ref{tab:sigcomppt1} and \ref{tab:sigcomppt2} for the Chinese and Dutch subsets, respectively. We compare the results against the techniques used by other competitors during SigComp 2011.

\begin{table}[h!]
\caption{SigComp'11 Chinese Results Comparison \citep{liwicki2011icdar}}
\setlength{\tabcolsep}{6pt}
    \centering
    \begin{tabular}{c c c c}
    \toprule
    \textbf{Models} & \textbf{Accuracy} & \textbf{FAR} & \textbf{EER} \\ \midrule
    Sabanci & 80.40\% & 19.62\%  & 20.315\% \\ \midrule
    Anonymous-1  & 73.10\% & 26.70\%  & 2.10\% \\ \midrule
    HDU  & 72.90\% & 26.98\%  & 27.24\% \\ \midrule
    DFKI  & 62.01\% & 38.15\%  & 38.24\% \\ \midrule
    VGG-16 \citep{Alvarez2016OfflineSV} & \textbf{88\%} & 8.2\%  & 13.2\% \\ \midrule
    Proposed Model  & 75.60\% & \textbf{7.60\%}  & \textbf{12.20\%} \\ \midrule
    \end{tabular}
    \label{tab:sigcomppt1}
\end{table}

\begin{table}[h!]
\caption{SigComp'11 Dutch Results Comparison \citep{liwicki2011icdar}}
\setlength{\tabcolsep}{6pt}
    \centering
    \begin{tabular}{c c c c}
    \toprule
    \textbf{Models} & \textbf{Accuracy} & \textbf{FAR} & \textbf{EER} \\ \midrule
    Sabanci & 82.91\% & 16.41\% & 17.17\% \\ \midrule
    Qatar  & \textbf{97.67\%} & \textbf{2.19\%} & \textbf{2.33\%} \\ \midrule
    HDU  & 87.80\% & 12.05\%  & 12.20\% \\ \midrule
    DFKI  & 75.84\% & 24.57\%  & 24.17\% \\ \midrule
    VGG-16 \citep{Alvarez2016OfflineSV} & 94\% & 13.32\%  & 8.22 \\ \midrule
    Proposed Model  & 76.83\% & 3.30\%  & 11.55\% \\ \midrule
    \end{tabular}
    \label{tab:sigcomppt2}
    
\end{table}

\subsection{SigWiComp 2013}
SigComp 2013 contains signatures from Japanese and Dutch writers in the same format as SigComp 2011. We only use the off-line Japanese signatures to test the multilingual signature verification capabilities of our architecture. The publicly available Task SigJapanese contains only off-line signatures and has a training set consisting of genuine and forged signatures from 11 distinct subjects. Each subject has 42 genuine and 36 forged signatures, resulting in a total of 858 signatures, of which 462 are genuine and 396 are forged. We divide the subjects into eight training and three testing subjects, with our model being trained on the genuine and forged signatures of eight subjects and evaluated on the remaining three subjects. This leaves us with 624 images for training and 234 images for testing.
\subsubsection{Training}
Our data pipeline preprocesses all signature images, and the resulting output is then used for model evaluation. The Relation Network training procedure is similar to that used for SigComp 2011, with 10,000 training episodes on genuine data and an additional 5000 training episodes for fine-tuning using forged and genuine samples.
\subsubsection{Results}
In the testing phase, we evaluate our fused architecture on the remaining three subjects by averaging the accuracies over 200 episodes. Each query image is compared with 1 or 5 genuine samples of the target class. The results are summarized in Table \ref{tab:sigwicomp}.

\begin{table}[h!]
\caption{SigWiComp'13 Japanese Results Comparison}
\setlength{\tabcolsep}{6pt}
    \centering
    \begin{tabular}{c c c c}
    \toprule
    \textbf{Models} & \textbf{Accuracy} & \textbf{FAR} & \textbf{EER} \\ \midrule
    Yilmaz \citep{yilmaz2011ijcb} & 90.72\% & 9.72\%  & 9.73\% \\ \midrule
    Kovari \citep{kovari2013elsevier}  & 72.70\% & 27.36\%  & 27.29 \\ \midrule
    Hassane \citep{hassan2012nips}  & 66.67\% & 33.33\% & 33.33\% \\ \midrule
    Djeddi \citep{djeddi2012springer}  & 72.10\% & 27.89\% & 27.90\% \\ \midrule
    Proposed Model  & \textbf{92\%} & \textbf{0.001\%} & \textbf{4\%} \\ \midrule
    \end{tabular}
    \label{tab:sigwicomp}
\end{table}

\section{Conclusion}
In conclusion, this study has demonstrated that our proposed approach, which combines hand-crafted features and deep feature embeddings, can produce results that are comparable to state-of-the-art models. Our fusion of the Relation Network training method with the OC-SVM algorithm has enabled us to implement a few-shot learning mechanism for biometric verification. By thresholding the relational score for genuine classes with a sigmoid-mapped distance from the hyperplane in the OC-SVM, we are able to strictly monitor the false acceptance rate, which is critical from a biometric standpoint.
The obtained results suggest that our approach can be used to build effective and efficient biometric verification systems. While there is still room for improvement, we believe that our approach is a simple and convenient solution that can be easily adopted in various applications. Future work could explore the potential of incorporating other deep learning techniques or alternative hand-crafted features to further improve the performance of our system.

\appendix
\label{app:theorem}


\vskip 0.2in
\bibliography{main}

\end{document}